\documentclass[sigconf]{acmart}
\usepackage[most]{tcolorbox}
\usepackage{xcolor}
\usepackage{booktabs}
\usepackage{multirow}
\usepackage[table]{xcolor} 
\usepackage{colortbl}
\usepackage{algorithm}
\usepackage{algorithmic}

\definecolor{promptgray}{RGB}{242,242,242}

\newtcolorbox{promptboxfull}[1]{
  enhanced,
  breakable,
  colback=promptgray,
  colframe=black,
  coltitle=white,
  colbacktitle=black,
  title={#1},
  fonttitle=\bfseries,
  boxrule=1pt,
  arc=2mm,
  left=3mm,
  right=3mm,
  top=2mm,
  bottom=2mm,
  toptitle=1.2mm,
  bottomtitle=1.2mm,
  lefttitle=3mm,
  righttitle=3mm,
  width=\linewidth,
  before skip=6pt,
  after skip=6pt
}

\AtBeginDocument{%
  }

\setcopyright{acmlicensed}
\copyrightyear{2018}
\acmYear{2018}
\acmDOI{XXXXXXX.XXXXXXX}
\acmConference[Conference acronym 'XX]{Make sure to enter the correct
  conference title from your rights confirmation email}{June 03--05,
  2018}{Woodstock, NY}

\acmISBN{978-1-4503-XXXX-X/2018/06}

\begin{document}

\title{A Multi-Agent Framework with Structured Reasoning and Reflective Refinement for Multimodal Empathetic Response Generation}

\author{Liping Wang\footnotemark[1]}
\email{lipingwang@mail.ustc.edu.cn}
\affiliation{%
  \institution{School of Information Science and Technology, USTC}
  \city{Hefei}
  \country{China}}

\author{Cheng Ye\footnotemark[1]}
\email{kyrieye@mail.ustc.edu.cn}
\affiliation{%
  \institution{School of Information Science and Technology, USTC}
  \city{Hefei}
  \country{China}}


\author{Weidong Chen\footnotemark[2]}
\email{chenweidong@ustc.edu.cn}
\affiliation{%
  \institution{School of Information Science and Technology, USTC}
  \city{Hefei}
  \country{China}
}

\author{Peipei Song}
\email{beta.songpp@gmail.com}
\affiliation{%
  \institution{School of Information Science and Technology, USTC}
  \city{Hefei}
  \country{China}}

\author{Bo Hu}
\email{hubo@ustc.edu.cn}
\affiliation{%
  \institution{School of Information Science and Technology, USTC}
  \city{Hefei}
  \country{China}}

\author{Zhendong Mao}
\email{zdmao@ustc.edu.cn}
\affiliation{%
  \institution{School of Information Science and Technology, USTC}
  \city{Hefei}
  \country{China}}

\renewcommand{\shortauthors}{Wang et al.}

\begin{abstract}
Multimodal empathetic response generation (MERG) aims to generate emotionally engaging and empathetic responses based on users' multimodal contexts. Existing approaches usually rely on an implicit one-pass generation paradigm from multimodal context to the final response, which overlooks two intrinsic characteristics of MERG: (1) Human perception of emotional cues is inherently structured rather than a direct mapping. The conventional paradigm neglects the hierarchical progression of emotion perception, leading to distorted emotional judgments. (2) Given the inherent complexity and ambiguity of human emotions, the conventional paradigm is prone to significant emotional biases, ultimately resulting in suboptimal empathy. In this paper, we propose a multi-agent framework for MERG, which enhances empathy through structured reasoning and reflective refinement. Specifically, we first introduce a structured empathetic reasoning-to-generation module that explicitly decomposes response generation via multimodal perception, consistency-aware emotion forecasting, pragmatic strategy planning, and strategy-guided response generation, providing a clearer intermediate path from multimodal evidence to response realization. Besides, we develop a global reflection and refinement module, in which a global reflection agent performs step-wise auditing over intermediate states and the generated response, eliminating existing emotional biases and empathy errors, and triggering targeted regeneration. Overall, such a closed-loop framework enables our model to gradually improve the accuracy of emotion perception and eliminate emotion biases during the iteration process. Experiments on several benchmarks, e.g., IEMOCAP and MELD, demonstrate that our model has superior empathic response generation capabilities compared to state-of-the-art methods.\footnote{Code will be released in the final version of the paper.}

\end{abstract}



\maketitle

\renewcommand{\thefootnote}{\fnsymbol{footnote}}
\footnotetext[1]{These authors contributed equally to this work.} 
\footnotetext[2]{Corresponding authors.}

\section{Introduction}

Recent advances in empathetic response generation have highlighted the need to produce responses that are not only fluent, but also supportive and contextually sensitive to the user’s emotional state \cite{rashkin2019empathetic, liu2021towards, deng2023knowledge}. In real interactions, however, such understanding rarely comes from language alone. Facial expressions, posture, and other nonverbal visual cues often provide important evidence about the speaker’s situation and affective condition. This has motivated growing interest in multimodal empathetic response generation (MERG), where response generation is grounded in both conversational context and nonverbal visual signals \cite{zhang2024stickerconv, zhang2025avamerg}. Compared with text-only settings, this task is substantially more challenging. The model must reason over heterogeneous modalities with nontrivial semantic gaps and cross-modal alignment difficulty \cite{zhang2020intraintermodal, hu2021mmgcn}. More importantly, effective empathy depends on both factual grounding in the situation and affective grounding in the speaker’s emotional experience. These two aspects are both essential for generating responses that are supportive and appropriate \cite{li2020empdg, li2022kemp, liu2021towards}.


Existing research has advanced empathetic response generation through contextual emotion modeling, knowledge-enhanced generation, multimodal understanding, and recent attempts at multimodal empathetic response generation \cite{li2020empdg, li2022kemp, zhang2020intraintermodal, hu2021mmgcn, zhang2024stickerconv}. Nevertheless, most of them still treat response generation as a largely implicit one-pass mapping from multimodal context to text without explicitly decomposing the intermediate reasoning process. Such a paradigm overlooks two fundamental characteristics of MERG: (1) Human emotional perception is inherently structured. For instance, when perceiving an image, humans sequentially identify affective elements, respond to these elements, and finally derive an emotional conclusion. In the one-pass generation diagram, the structured reasoning process remains implicit, making it difficult to localize and correct early-stage errors, especially when textual and visual cues are only partially aligned, which severely compromises both the accuracy and interpretability of emotional perception. (2) Furthermore, since human emotions possess intrinsic complexity and ambiguity, emotion resonance requires a progressive evolutionary process. One-pass generation diagram oversimplifies this dynamic and is prone to significant affective errors. These errors may propagate to downstream generation, leading to responses that are fluent but insufficiently grounded, strategically misaligned, or only superficially empathetic.

\begin{figure}[h]
    \centering
    \includegraphics[width=1\linewidth]{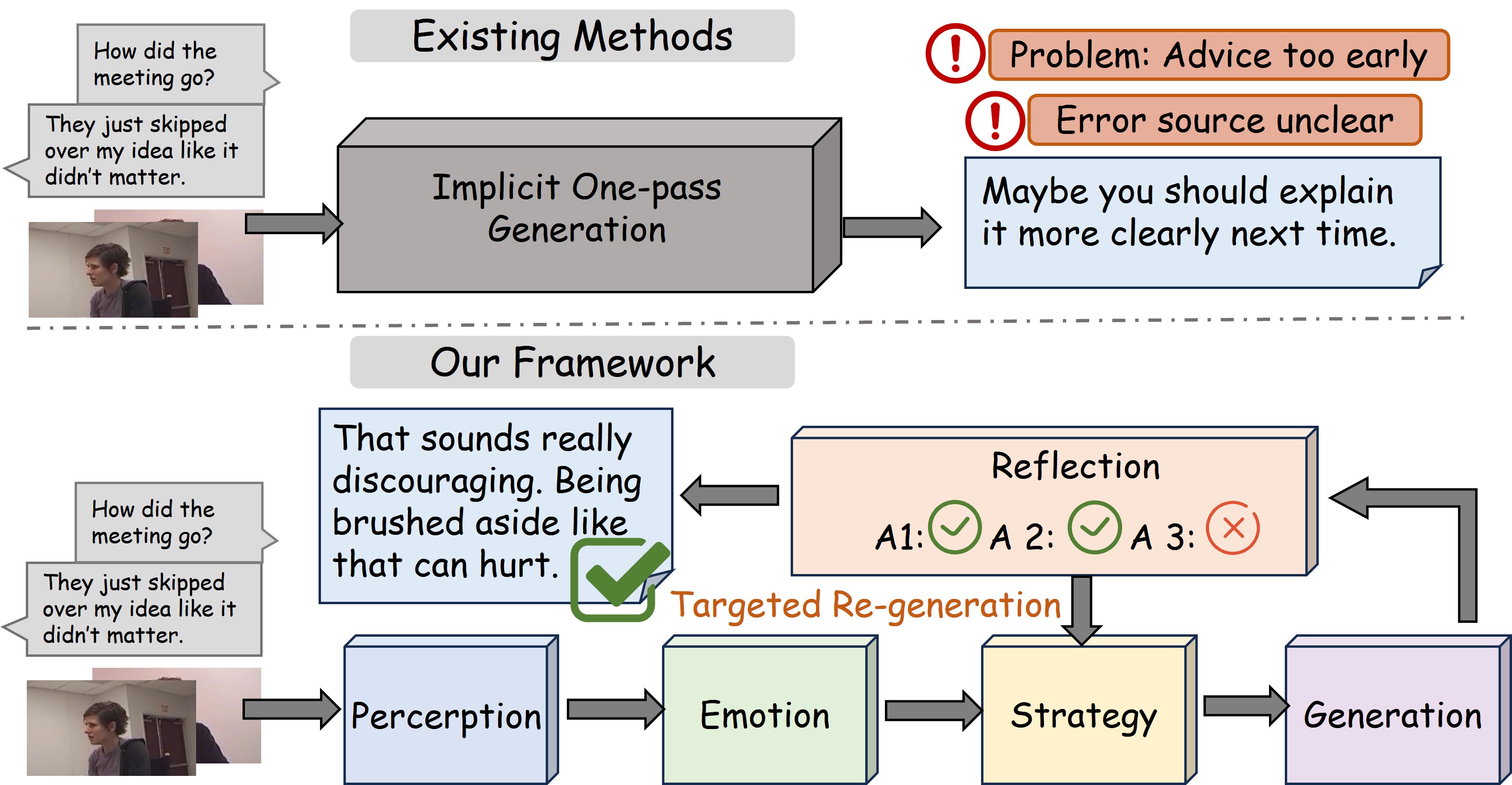}
    \caption{Conventional methods generate responses in a single implicit pass, making errors difficult to trace. Our framework instead adopts structured empathetic reasoning and reflective refinement, enabling targeted re-generation and improved final response selection.}
    \label{fig:first}
\end{figure}

To address these limitations, we propose a multi-agent framework for multimodal empathetic response generation, consisting of a structured empathetic reasoning-to-generation pipeline and a global reflection and refinement module. Specifically, the structured empathetic reasoning-to-generation pipeline decomposes generation into four coordinated stages. Firstly, we extract concrete visual and textual evidence for emotion perception from multimodal input. Subsequently, we design a consistency-aware emotion forecaster to predict the emotion conclusion for the response through focus on the consistency between extracted emotion evidence. Besides, we introduce a pragmatic strategy planning agent to determine how the response should be delivered. Finally, we generate the empathetic response with the guidance of the aggregation for above information. Overall, this pipeline provides a clearer and more controllable path from multimodal evidence to empathetic response generation. Building on this process, we design a global reflection and refinement module. A global reflection agent is designed to step-wise audit over the intermediate outputs and the generated response, identify the earliest faulty stage, and trigger targeted re-generation. Besides, instead of simply returning the last refined response, the framework selects the final output from the refinement history, which helps reduce error propagation and over-correction. Finally, through the closed-loop combination of progressive structured reasoning-to-generation and global reflection and refinement, our framework improves the consistency, strategic coherence, and emotional appropriateness of the generated response. In summary, our main contributions are as follows:

$\bullet$ We propose a multi-agent framework for multimodal empathetic response generation. Specifically, we design a closed-loop iterative generation framework that continuously optimizes the responses through a reflection and refinement mechanism, which enhances the ability to produce empathetic responses that are emotionally appropriate, strategically coherent, and well supported by multimodal context. 

$\bullet$ We introduce a structured reasoning-to-generation module that explicitly decomposes response generation via four dense components, providing a clearer intermediate path from multimodal evidence to response realization. Besides, we develop a global reflection and refinement module, in which a global reflection agent performs step-wise auditing over intermediate states and the generated response, eliminating existing emotional biases and empathy errors, and triggering targeted regeneration.
    
$\bullet$ Extensive experiments on several benchmarks (\emph{i.e.,} IEMOCAP and MELD) demonstrate the effectiveness of the proposed method and two core components,  achieving relative improvements of 16.4\%/16.0\% in Acc./Empathy on the IEMOCAP dataset.

\section{Related Work}

Prior work relevant to our study mainly falls into three lines: multimodal emotion understanding, empathetic response generation and emotional support conversation, and LLM-based multi-agent refinement. These lines respectively relate to the two core components of our framework: the structured empathetic reasoning-to-generation pipeline, which requires grounded multimodal perception, emotion forecasting, strategy planning, and response realization, and the global reflection and refinement module, which performs targeted revision over intermediate decisions and generated responses.

\subsection{Multimodal Emotion Analysis}

Reliable affect understanding is a prerequisite for multimodal empathetic response generation. Early work such as DialogueRNN~\cite{majumder2019dialoguernn} modeled speaker-aware emotional dynamics in conversation, while MELD~\cite{poria2019meld} established a widely used benchmark with synchronized textual, acoustic, and visual streams. Subsequent studies moved beyond simple multimodal fusion toward more explicit modeling of conversational structure and cross-modal interaction. Zhang et al.~\cite{zhang2020intraintermodal} characterized both intra-modal and inter-modal influences in conversational emotion recognition, and MMGCN~\cite{hu2021mmgcn} modeled cross-modal and inter-speaker dependencies through graph-based reasoning. Later work further strengthened robustness and reliability under challenging conversational conditions. M3Net~\cite{chen2023m3net}, for instance, revisited multimodal graph reasoning from multivariate and multi-frequency perspectives, while CMERC~\cite{tu2024cmerc} highlighted the importance of calibrated prediction under noisy or ambiguous evidence. These studies provide an important basis for multimodal affect understanding. However, they mainly treat emotion recognition as the final objective, rather than as an intermediate reasoning stage that must remain consistent with downstream strategy planning and empathetic response generation.

\subsection{Empathetic Response Generation}

Empathetic response generation aims to produce responses that are emotionally appropriate and contextually supportive to the speaker. EmpatheticDialogues~\cite{rashkin2019empathetic} established a widely adopted benchmark for this task, and subsequent studies have mainly improved generation quality through stronger affect modeling, external knowledge integration, and more explicit intermediate reasoning. Early methods focused on enhancing emotional awareness in generation. EmpDG~\cite{li2020empdg} incorporated dialogue-level and token-level emotion signals, while MIME~\cite{majumder2020mime} modeled emotional mimicry to improve empathetic expression. Another line of work introduced external knowledge to better support response generation. KEMP~\cite{li2022kemp} integrated commonsense and emotional lexical knowledge through an emotional context graph, and SEEK~\cite{wang2022seek} modeled emotion flow together with emotion--knowledge interaction. More recent work has begun to make the generation process more explicit. CASE \cite{zhou2023case} aligned cognition and affection in a coarse-to-fine manner, and IAMM \cite{yang2024iamm} improved response generation with iterative associative memory. Related progress in emotional support dialogue further highlights the importance of explicit intermediate structure, such as decoupled strategy prediction in EmoDynamiX \cite{wan2025emodynamix}. On the multimodal side, STICKERCONV \cite{zhang2024stickerconv} introduced multimodal empathetic responses with stickers, AvaMERG \cite{zhang2025avamerg} extended the task to synchronized text, speech, and facial video, and personality-aware multimodal empathy modeling \cite{wu2025traits} incorporated multimodal user cues together with personal traits. Recent work further advances this direction through open-source avatar-based systems such as EmpathyEar \cite{fei2024empathyear} and community-level task development such as the Ava-MERG Challenge \cite{zhang2025avamerg_challenge}, while E3RG \cite{lin2025e3rg} explores explicit modular decomposition for multimodal empathetic response generation. Despite this progress, most existing methods either remain largely end-to-end or expose only part of the intermediate generation process. Few frameworks explicitly organize multimodal perception, emotion forecasting, pragmatic strategy planning, and response generation within a unified pipeline.

\subsection{LLM-based Multi-Agent Systems and Iterative Refinement}
Recent advances in LLM-based agentic frameworks suggest a natural way to decompose complex generation tasks into specialized roles. Representative systems such as AutoGen~\cite{wu2024autogen} and MetaGPT~\cite{hong2024metagpt} demonstrate that structured role assignment and workflow design can improve coordination over multi-step tasks. In parallel, iterative refinement methods such as Reflexion~\cite{shinn2023reflexion} and Self-Refine~\cite{madaan2023selfrefine} show that linguistic feedback can improve generation quality without parameter updates. In the dialogue domain, MultiAgentESC~\cite{xu2025multiagentesc} further illustrates the promise of collaborative LLM agents for emotional support conversation. Nevertheless, existing agentic and reflection-based methods are still predominantly text-centric, and are rarely designed for multimodal empathetic response generation. More importantly, they usually do not provide a structured mechanism to audit intermediate states step by step, identify the earliest faulty stage, and perform targeted downstream re-generation across perception, emotion forecasting, strategy planning, and response realization. Recent multimodal systems such as E3RG \cite{lin2025e3rg} also adopt explicit decomposition for empathetic response generation, but they do not provide a step-wise auditing mechanism that identifies and corrects errors at the earliest responsible stage. Our framework builds on this line of research, but adapts it to multimodal empathetic response generation through a structured reasoning-to-generation pipeline coupled with a global reflection and  refinement module.

\section{Method}

\subsection{Task Formulation and Framework Overview}

Formally, the multimodal empathetic response generation task is defined as follows. Let $\mathcal{C} = \{(s_i, u_i)\}_{i=1}^{N}$ denote the conversational history, where $s_i$ and $u_i$ represent the speaker and the textual utterance at turn $i$, respectively. Paired with $\mathcal{C}$ is a sequence of raw video clips $\mathcal{V} = \{v_i\}_{i=1}^{N}$. To align the continuous visual stream with the discrete textual context, we extract a set of representative keyframes $\tilde{\mathcal{V}}$ from $\mathcal{V}$. Ultimately, our objective is to generate an empathetic and pragmatically appropriate next-turn response $R$ conditioned on the multimodal context $(\mathcal{C}, \tilde{\mathcal{V}})$.

\begin{figure*}[t]
    \centering
    \includegraphics[width=0.8\linewidth]{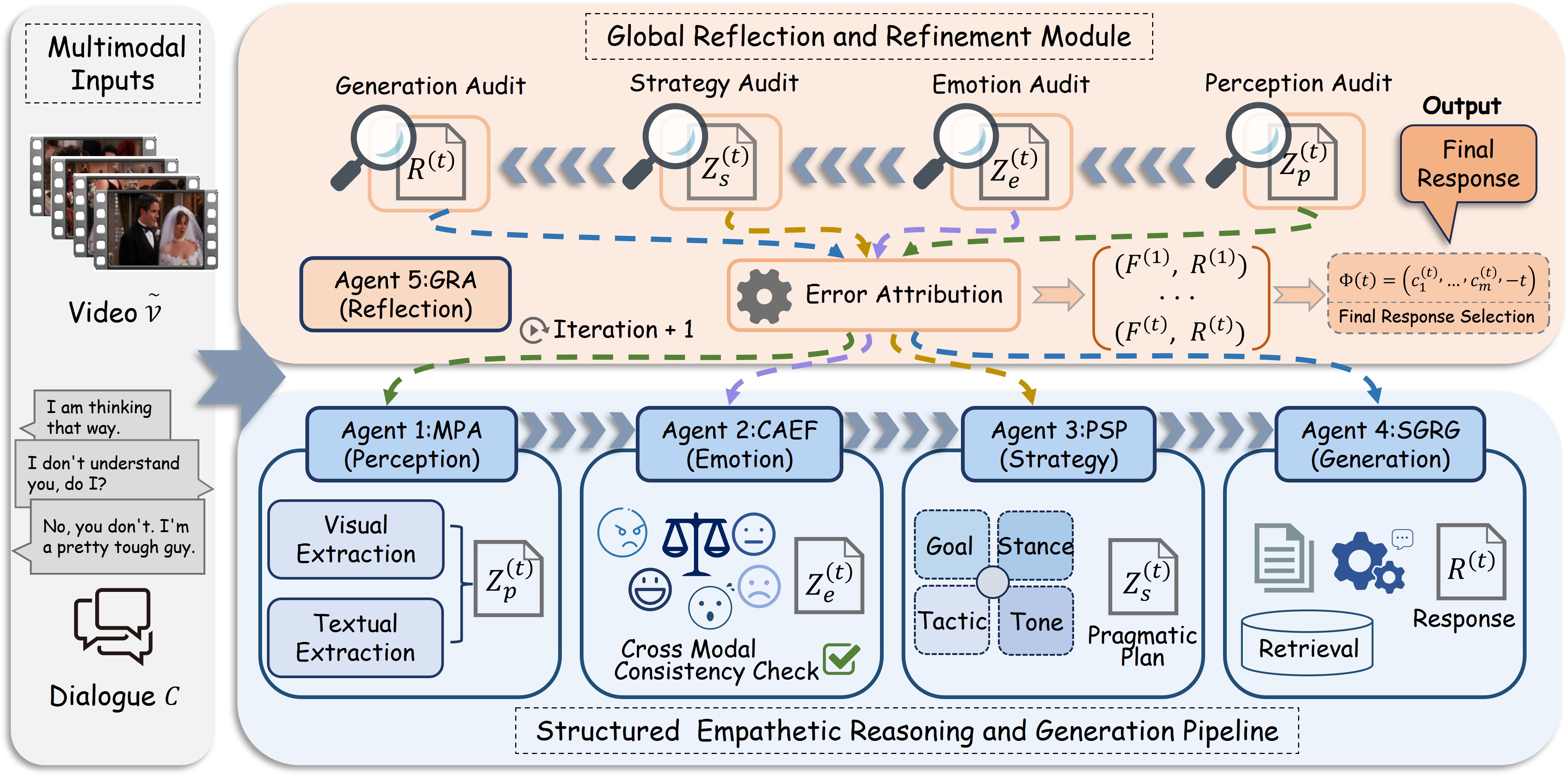}
    \caption{Overview of the proposed framework. Given the multimodal context, the structured empathetic reasoning-to-generation pipeline progressively performs multimodal perception, emotion forecasting, strategy planning, and response generation. A global reflection and refinement module then audits intermediate states and the generated response, attributes errors to the responsible stage, and triggers targeted re-generation. The final response is selected from the refinement history.}
    \label{fig:framework}
\end{figure*}

We propose a multi-agent framework for multimodal empathetic response generation, composed of a structured empathetic reasoning-to-generation pipeline and a global reflection and refinement module. The former progressively conducts multimodal perception, emotion forecasting, strategy planning, and response generation, while the latter inspects the intermediate reasoning process and the final response to provide targeted feedback for iterative refinement. An overview of the framework is shown in Fig.~\ref{fig:framework}.

\subsection{Structured Reasoning-to-Generation}

The first four agents form a structured empathetic reasoning-to-generation pipeline. We adopt this progressive design because multimodal empathetic response generation involves a sequence of tightly coupled yet distinct processes, rather than a direct mapping from context to the final response~\cite{zhang2025towards,xu2025multiagentesc}. By explicitly organizing multimodal perception, emotion forecasting, strategy planning, and response generation into a structured empathetic reasoning-to-generation pipeline, the framework establishes a clearer intermediate path toward the final response and provides a more controllable basis for subsequent reflection and refinement.

\subsubsection{\textbf{Multimodal Perception Agent}}

Multimodal Perception Agent (MPA) is responsible for extracting concrete visual and textual evidence from the raw multimodal context. Since multimodal empathetic response generation depends on both verbal and nonverbal cues, organizing the input into explicit perceptual evidence provides a cleaner basis for downstream reasoning~\cite{zhang2025towards}. Acting as the observation stage of the pipeline, it identifies response-relevant cues from the dialogue history $\mathcal{C}$ and sampled video frames $\tilde{\mathcal{V}}$. The process is formulated as
\begin{equation}
    Z_p = \mathrm{MPA}(\mathcal{C}, \tilde{\mathcal{V}})
\end{equation}
where $Z_p$ comprises the grounded multimodal evidence for downstream emotion forecasting and strategy planning.


\subsubsection{\textbf{Consistency-Aware Emotion Forecaster}}

The Consistency-Aware Emotion Forecaster (CAEF) is designed to predict the emotion for the final response. Because textual and visual signals may provide complementary or partially inconsistent cues, the emotion forecasting agent jointly models the dialogue context, visual observations, and grounded perceptual evidence to infer the emotion that should guide subsequent planning and generation~\cite{hazarika2020misa}. Formally,
\begin{equation}
    Z_e = \mathrm{CAEF}(\mathcal{C}, \tilde{\mathcal{V}}, Z_p)
\end{equation}
where $Z_e$ denotes the forecasted emotion representation.


\subsubsection{\textbf{Pragmatic Strategy Planner}}
The Pragmatic Strategy Planner (PSP) is responsible for determining how the final response should be delivered. This agent is introduced because the forecasted emotion alone is insufficient to specify communicative intention and expression style~\cite{liu2021towards, cheng-etal-2022-improving}. Based on the dialogue context, perceptual evidence, and forecasted emotion, the PSP constructs a structured pragmatic plan with four dimensions: communication goal ($\gamma$), interpersonal stance ($\alpha$), pragmatic tactic ($\rho$), and linguistic tone ($\tau$).  By explicitly defining $Z_s = \{\gamma, \alpha, \rho, \tau\}$,the framework converts high-level empathetic intent into a structured pragmatic plan. Formally,
\begin{equation}
    Z_s = \mathrm{PSP}(\mathcal{C}, Z_p, Z_e)
\end{equation}
where $Z_s$ denotes the pragmatic plan for the subsequent response generation agent.



\subsubsection{\textbf{Strategy-Grounded Response Generator}}

The Strategy-Grounded Response Generator (SGRG) is responsible for generating the final response conditioned on the dialogue context and the outputs \(Z_p\), \(Z_e\), and \(Z_s\) from the preceding agents. Specifically, it realizes the pragmatic plan in natural language while incorporating the perceptual evidence, forecasted emotion, and retrieved auxiliary context from $ \mathcal{K}\ $. The generation process is formulated as:
\begin{equation}
    R = \mathrm{SGRG}(\mathcal{C}, \tilde{\mathcal{V}},  Z_p, Z_e, Z_s,  \mathcal{K})
\end{equation}
where $R$ denotes the generated response.


\subsection{Global Reflection and Refinement}

The Global Reflection Agent (GRA) ensures global alignment through an iterative refinement process. Let \(t \in \{1, \dots, T\}\) denote the refinement iteration index, where \(t=1\) corresponds to the initial output of the structured empathetic reasoning-to-generation pipeline. At each iteration, the GRA performs two steps: (1) Step-wise Functional Audit and (2) Error Attribution and Pipeline Re-generation. After refinement terminates, a deterministic evaluation function selects the final response from the recorded generation history.

\textbf{1. Step-wise Functional Audit:} In each iteration $t$, the GRA performs a sequential check to verify whether each agent in the  structured empathetic reasoning-to-generation pipeline has fulfilled its specific role. It evaluates: 
(i) the MPA ($Z_p^{(t)}$) for factual accuracy and multimodal interpretation; 
(ii) the CAEF ($Z_e^{(t)}$) for the plausibility of the forecasted emotion under multimodal evidence; 
(iii) the PSP ($Z_s^{(t)}$) for the appropriateness of the strategic plan; 
and (iv) the SGRG for whether the generated utterance $R^{(t)}$ complies with the constraints set by $Z_p, Z_e,$ and $Z_s$.

\textbf{2. Error Attribution and Pipeline Re-generation:} If a mismatch is detected, the GRA identifies the specific agent responsible for the error. To prevent error propagation, the structured empathetic reasoning-to-generation pipeline is re-activated starting from the failed agent. For example, if the error is localized to the CAEF ($Z_e$), the feedback $F^{(t)}$ is used to re-generate a corrected emotional state $Z_e^{(t+1)}$. This update automatically triggers new rounds of strategic planning ($Z_s^{(t+1)}$) and response generation ($R^{(t+1)}$).

\begin{equation}
F^{(t)} = \mathrm{GRA}(\mathcal{C}, \tilde{\mathcal{V}}, Z_p^{(t)}, Z_e^{(t)}, Z_s^{(t)}, R^{(t)})
\end{equation}
\begin{equation}
\{Z_k^{(t+1)}, \dots, R^{(t+1)}\} = \mathrm{Pipeline}(\mathcal{C}, \tilde{\mathcal{V}}, Z^{(t)}, \pi_k(F^{(t)}))
\end{equation}
where \(k \in \{p, e, s, r\}\) denotes the agent identified as the source of error, \(\pi_k(F^{(t)})\) denotes its correction instruction, and \(Z^{(t)}=\{Z_p^{(t)}, Z_e^{(t)}, Z_s^{(t)}, R^{(t)}\}\) denotes the outputs from iteration \(t\), which are retained to support feedback-guided, targeted refinement in the next iteration. The closed-loop continues until all functional checks are passed or the maximum iteration count is reached.


\textbf{3. Final Response:} To mitigate semantic drift and over-correction in prolonged iterations, the framework selects the optimal response $R_{\mathrm{final}}$ from the history $\mathcal{H}=\{(R^{(t)},F^{(t)})\}_{t=1}^{T}$ rather than defaulting to $R^{(T)}$ \cite{huang2024large}. We map each audit feedback $F^{(t)}$ into a lexicographic evaluation tuple:

\begin{equation}
    \Phi(t) = \left( c_1^{(t)}, c_2^{(t)}, \dots, c_m^{(t)}, -t \right)
\end{equation}
where $c_i^{(t)} \in \{0, 1\}$ are binary check results extracted from $F^{(t)}$, ordered by architectural priority. The term $-t$ serves as a regularization penalty to favor earlier states during ties. The final output is then determined by $R_{final} = R^{(t^*)}$, where the optimal index $t^* = \arg\max_{t \in \{1, \dots, T\}}^{\text{lex}} \Phi(t)$ ensures the selection of the most consistent and least over-optimized generation.

\begin{algorithm}[t]
\caption{Overall procedure of the proposed framework.}
\label{alg:drr}
\begin{algorithmic}[1]
\REQUIRE Multimodal context $(\mathcal{C}, \tilde{\mathcal{V}})$, retrieval memory $\mathcal{K}$; max iterations $T_{\max}$.
\ENSURE Final refined response $R^*$.

\STATE \textbf{// Stage 1: Structured Reasoning-to-Generation Pipeline }
\STATE $Z_p^{(1)} \leftarrow \mathrm{MPA}(\mathcal{C}, \tilde{\mathcal{V}})$
\STATE $Z_e^{(1)} \leftarrow \mathrm{CAEF}(\mathcal{C}, \tilde{\mathcal{V}}, Z_p^{(1)})$
\STATE $Z_s^{(1)} \leftarrow \mathrm{PSP}(\mathcal{C}, Z_p^{(1)}, Z_e^{(1)})$ \COMMENT{$Z_s = \{\gamma, \alpha, \rho, \tau\}$}
\STATE $R^{(1)} \leftarrow \mathrm{SGRG}(\mathcal{C}, \tilde{\mathcal{V}}, Z_p^{(1)}, Z_e^{(1)}, Z_s^{(1)}, \mathcal{K})$
\STATE $R^* \leftarrow R^{(1)}$
\STATE $F^{(1)} \leftarrow \mathrm{GRA}(\mathcal{C}, \tilde{\mathcal{V}}, Z_p^{(1)}, Z_e^{(1)}, Z_s^{(1)}, R^{(1)})$
\STATE $\mathcal{H} \leftarrow \{(R^{(1)}, F^{(1)})\}$

\FOR{$t = 1$ \TO $T_{\max}$}
    \IF{$F^{(t)}.\mathtt{is\_valid} = \mathtt{True}$}
        \STATE \textbf{break}
    \ENDIF
    
    \STATE $ F_k^{(t)} \leftarrow \pi_k(F^{(t)}), \quad k \in \{p,e,s,r\}$ is the identified error source

    \STATE \textbf{// Stage 2: Global Reflection and Refinement Module}
    \FORALL{components $i$ before $k$}
        \STATE $Z_i^{(t+1)} \leftarrow Z_i^{(t)}$ \COMMENT{Keep previously validated outputs}
    \ENDFOR
    \STATE Re-run the pipeline from component $k$ with corrective feedback $F_k^{(t)}$
    \STATE $\{Z_k^{(t+1)}, \dots, R^{(t+1)}\} \leftarrow \mathrm{Pipeline}(\mathcal{C}, \tilde{\mathcal{V}}, \mathcal{K}, Z^{(t)}, F_k^{(t)})$
    \STATE $F^{(t+1)} \leftarrow \mathrm{GRA}(\mathcal{C}, \tilde{\mathcal{V}}, Z_p^{(t+1)}, Z_e^{(t+1)}, Z_s^{(t+1)}, R^{(t+1)})$
    \STATE $\mathcal{H} \leftarrow \mathcal{H} \cup \{(R^{(t+1)}, F^{(t+1)})\}$
\ENDFOR

\STATE \textbf{// Stage 3: Final Response}
\STATE $t^* \leftarrow \arg\max_t^{\mathrm{lex}} \Phi(t)$
\STATE $R^* \leftarrow R^{(t^*)}$
\RETURN $R^*$
\end{algorithmic}
\end{algorithm}

\section{Experiment}

We evaluate the proposed framework on IEMOCAP \cite{busso2008iemocap} and MELD \cite{poria2019meld} through automatic evaluation, human evaluation. In addition to overall comparison results, we further analyze the contribution of key components and provide qualitative case studies.

\subsection{Experimental Setup}

\subsubsection{Datasets}

We conduct experiments on two widely used multimodal dialogue datasets, IEMOCAP and MELD. \textbf{IEMOCAP} is a multimodal dyadic conversation dataset with roughly 12 hours of audiovisual recordings, accompanied by transcripts and emotion labels \cite{busso2008iemocap}. Following prior settings \cite{mao2021dialoguetrm,wu2025ergm}, we use Session 5 for evaluation. \textbf{MELD} is a multimodal multi-party dialogue benchmark built from the TV series Friends. It includes 1,433 dialogues and about 13K utterances annotated with emotion and sentiment labels \cite{poria2019meld}. We adopt the official split and report results on its test set. In both datasets, the model takes the dialogue history and sampled video keyframes as input.

\subsubsection{Metrics}

We evaluate our model using both automatic metrics and human evaluation. Following prior work, we report Perplexity (PPL), Emotion Accuracy (Acc.), Distinct-1/2, and BERTScore \cite{li2020empdg,zhao2023empsoa,wu2025ergm} for automatic evaluation. PPL evaluates the fluency and overall generation difficulty of the produced response under the language model, where a lower value indicates that the response is more natural and better formed. Acc. measures the correctness of target emotion prediction. Distinct-1/2 measures lexical diversity based on the proportions of distinct unigrams and bigrams in generated responses \cite{li2016diversity}. We also report the precision, recall, and F1 of BERTScore to evaluate semantic similarity between generated and reference responses at the contextual embedding level \cite{zhang2020bertscore}. Lower PPL indicates better performance, while higher Acc., Distinct-1/2, and BERTScore indicate stronger results.
In addition to automatic metrics, we perform human evaluation to provide a complementary assessment of response quality. The generated responses are rated on a 5-point Likert scale from three aspects. \textbf{Empathy} measures whether the response appropriately reflects understanding of the speaker’s emotional state and situation. \textbf{Coherence} evaluates whether the response is consistent with the dialogue context and communicative focus. \textbf{Fluency} assesses the naturalness, readability, and grammatical well-formedness of the response.

\subsubsection{Implementation Details}

All five agents in our framework are instantiated with the same backbone model, \textbf{Qwen3.5:27B}, and are served locally through \textbf{Ollama}\footnote{\url{https://ollama.com}} \cite{qwen2026qwen3527b,ollama2026}. The proposed framework is fully \textbf{training-free}, with all results obtained by inference only under the original model parameters. For evaluation, we follow standard benchmark settings, using Session 5 of IEMOCAP for testing \cite{mao2021dialoguetrm,wu2025ergm} and the official test split of MELD \cite{poria2019meld}. In the closed-loop reflection module, we allow at most two refinement iterations after the initial pipeline pass, which provides a practical balance between computational cost and over-correction. When retrieval is enabled, the memory is built prior to evaluation to prevent information leakage, using Session 2 for IEMOCAP and the development split for MELD. Unless otherwise stated, all experiments are conducted on a single NVIDIA RTX 3090 GPU.

\subsection{Main Results}

We first examine whether the proposed framework improves overall response generation quality under both automatic and human evaluation. Table~\ref{tab:main_results} compares our method with representative baselines on IEMOCAP and MELD.

\begin{table*}[t]
\centering
\footnotesize
\setlength{\tabcolsep}{3.6pt}
\renewcommand{\arraystretch}{0.9} 
\setlength{\aboverulesep}{0pt}
\setlength{\belowrulesep}{0pt}
\caption{Comparison with baselines on IEMOCAP and MELD under automatic and human evaluation metrics. The best and second-best results are marked with bold and underline, respectively.}
\label{tab:main_results}
\resizebox{\textwidth}{!}{%
\begin{tabular}{llcccccccccc}
\toprule
\multirow{2}{*}{Datasets} & \multirow{2}{*}{Methods} & \multicolumn{7}{c}{Automatic Evaluation} & \multicolumn{3}{c}{Human Evaluation} \\
\cmidrule(lr){3-9} \cmidrule(lr){10-12}
& & PPL & Dist-1 & Dist-2 & Acc. & $P_{\text{BERT}}$ & $R_{\text{BERT}}$ & $F_{\text{BERT}}$ & Emp. & Coh. & Flu. \\
\midrule

\multirow{15}{*}{IEMOCAP}
& \multicolumn{11}{l}{\textit{Non-LLM methods}} \\
& MoEL \cite{lin2019moel} & 48.34 & 0.88 & 3.56 & 64.18 & 80.11 & 82.92 & 81.49 & 2.91 & 3.09 & 3.37 \\
& MIME \cite{majumder2020mime} & 44.81 & 0.92 & 3.67 & 65.52 & 79.93 & 82.17 & 81.03 & 2.88 & 3.14 & 3.34 \\
& EmpDG \cite{li2020empdg} & 43.68 & 0.96 & 3.89 & 64.27 & 80.33 & 83.75 & 82.00 & 2.94 & 3.22 & 3.42 \\
& CEM \cite{sabour2022cem} & 45.03 & 1.09 & 4.15 & 64.83 & 80.96 & 83.83 & 82.37 & 3.02 & 3.27 & 3.65 \\
& SEEK \cite{wang2022seek} & 44.35 & 1.03 & 3.94 & 66.40 & 80.71 & 83.34 & 82.01 & 3.04 & 3.24 & 3.58 \\
& EmpSOA \cite{zhao2023empsoa}& 38.17 & 1.16 & 4.72 & 67.35 & 81.25 & 84.40 & 82.79 & 3.06 & 3.28 & 3.61 \\
& CASE \cite{zhou2023case} & 36.59 & 1.18 & 4.93 & 65.94 & 81.04 & 83.89 & 82.44 & 3.05 & 3.25 & 3.63 \\
& IAMM \cite{yang2024iamm} & 38.36 & 1.15 & 4.69 & 66.78 & 81.45 & 84.81 & 83.10 & 3.07 & 3.26 & 3.57 \\
& ERGM \cite{wu2025ergm} & 30.21 & 3.16 & 11.18 & \underline{72.05} & \underline{83.57} & \underline{86.33} & \underline{84.93} & \underline{3.31} & 3.44 & 3.80 \\
\cmidrule(lr){2-12}
\rowcolor{gray!20} & \multicolumn{11}{l}{\textit{LLM-based methods}} \\
\rowcolor{gray!20} & GPT-4-V (+ 5-shot) & - & 3.89 & \underline{26.06} & 58.74 & 80.11 & 82.56 & 81.31 & 3.16 & \underline{3.58} & \underline{4.66} \\
\rowcolor{gray!20} & Llama 3.2-11B & \underline{7.62} & \underline{3.95} & 22.08 & 60.52 & 82.31 & 82.06 & 82.18 & \underline{3.31} & 3.44 & 4.59 \\
\cmidrule(lr){2-12}
\rowcolor{gray!20} & \multicolumn{11}{l}{\textit{MultiAgent methods}} \\
\rowcolor{gray!20} & Ours & \textbf{1.94} & \textbf{37.25} & \textbf{81.10} & \textbf{83.87} & \textbf{84.03} & \textbf{86.64} & \textbf{85.28} & \textbf{3.84} & \textbf{3.96} & \textbf{4.91} \\
\midrule

\multirow{15}{*}{MELD}
& \multicolumn{11}{l}{\textit{Non-LLM methods}} \\
& MoEL \cite{lin2019moel} & 56.84 & 0.71 & 3.22 & 57.93 & 80.98 & 81.75 & 81.36 & 2.90 & 3.16 & 3.29 \\
& MIME \cite{majumder2020mime} & 48.50 & 0.64 & 2.88 & 56.90 & 79.50 & 80.29 & 79.89 & 2.98 & 3.22 & 3.36 \\
& EmpDG \cite{li2020empdg} & 50.51 & 0.89 & 4.05 & 57.62 & 80.20 & 81.59 & 80.88 & 2.92 & 3.23 & 3.26 \\
& CEM \cite{sabour2022cem} & 54.00 & 0.97 & 4.36 & 57.55 & 78.77 & 80.40 & 79.57 & 2.95 & 3.20 & 3.30 \\
& SEEK \cite{wang2022seek} & 54.72 & 1.01 & 4.54 & 58.95 & 81.33 & 82.97 & 82.14 & 2.97 & 3.27 & 3.33 \\
& EmpSOA \cite{zhao2023empsoa} & 53.33 & 1.02 & 4.60 & 59.69 & 82.71 & \underline{84.68} & 83.68 & 2.96 & 3.27 & 3.38 \\
& CASE \cite{zhou2023case} & 36.02 & 1.10 & 5.07 & 60.48 & 81.59 & 83.91 & 82.73 & 2.92 & 3.25 & 3.42 \\
& IAMM \cite{yang2024iamm} & 36.02 & 1.08 & 4.85 & 58.72 & 82.38 & 84.55 & 83.45 & 2.95 & 3.24 & 3.47 \\
& ERGM \cite{wu2025ergm} & 29.61 & 2.96 & 10.84 & \textbf{66.45} & \underline{83.95} & \textbf{85.28} & \underline{84.61} & \underline{3.38} & 3.47 & 3.85 \\
\cmidrule(lr){2-12}
\rowcolor{gray!20} & \multicolumn{11}{l}{\textit{LLM-based methods}} \\
\rowcolor{gray!20} & GPT-4-V (+ 5-shot) & - & \underline{3.92} & \underline{25.58} & 54.62 & 78.51 & 75.96 & 77.21 & 3.18 & \underline{3.55} & \underline{4.59} \\
\rowcolor{gray!20} & Llama 3.2-11B & \underline{7.36} & 3.86 & 21.03 & 55.18 & 81.39 & 80.64 & 81.01 & 3.30 & 3.46 & 4.56 \\
\cmidrule(lr){2-12}
\rowcolor{gray!20} & \multicolumn{11}{l}{\textit{MultiAgent methods}} \\
\rowcolor{gray!20} & Ours & \textbf{2.06} & \textbf{28.14} & \textbf{77.04} & \underline{64.79} & \textbf{85.30} & 84.23 & \textbf{84.72} & \textbf{3.75} & \textbf{3.85} & \textbf{4.88} \\
\bottomrule
\end{tabular}%
}
\end{table*}

\textbf{Automatic Evaluation.}
Under automatic evaluation, our method shows a favorable overall trend on both datasets. On IEMOCAP, it improves PPL, Dist-1/2, Acc., and BERT-F1 over the compared baselines. On MELD, it also maintains competitive results on PPL, Dist-1/2, and BERT-F1, while remaining slightly below ERGM on Acc. Overall, these results suggest that the proposed framework is particularly helpful for improving final response generation quality, while its benefit on target-emotion forecasting is somewhat more sensitive to the dataset.

The lower PPL suggests that our model makes response generation more stable and better constrained, since the response is produced from a structured reasoning process with grounded perception, emotion forecasting, pragmatic planning, and subsequent reflection, rather than from raw multimodal input alone. The much higher Dist-1/2 further indicates that the framework is less prone to generic and repetitive responses. A likely reason is that explicit strategy planning provides more diverse communicative intentions before surface realization, while the reflection module further helps correct underspecified outputs. As a result, the generated responses become more diverse without an obvious loss of semantic relevance, which is also consistent with the BERTScore results.

\textbf{Human Evaluation.}
The human evaluation results show a similar overall trend. Our method obtains consistently favorable scores on empathy, coherence, and fluency across both datasets, suggesting that its improvements are not limited to automatic metrics. This tendency is broadly consistent with the framework design: multimodal evidence extraction, emotion forecasting, and strategy planning are explicitly separated in the pipeline, and stage-level mismatches can be further corrected through reflection before final response selection.

\subsection{Component and Design Analysis}
We next analyze two questions: whether each agent in the proposed framework is necessary, and whether the cascaded reasoning-to-generation pipeline is preferable to an independent organization. The corresponding results are reported in Table~\ref{tab:component_analysis}.

\textbf{Effect of Individual Agents (MPA, CAEF, PSP, and GRA).}
We first perform agent-wise ablation to assess the role of each core agent in the proposed framework. Removing any core agent leads to degradation on at least part of the metrics, suggesting that all four agents make meaningful contributions. In particular, removing MPA or GRA causes the clearest drop in emotion accuracy, underscoring the role of grounded multimodal evidence and reflective correction. Removing PSP also weakens diversity and semantic matching, indicating that explicit pragmatic planning helps bridge emotion understanding and final response generation. Although removing CAEF yields a lower PPL, it also reduces diversity and BERTScore, suggesting that lower perplexity alone does not necessarily imply better empathetic response quality.

\textbf{Effect of Cascaded reasoning-to-generation Pipeline Organization.}
We further compare the proposed cascaded reasoning-to-generation pipeline with an independent variant to examine the role of explicit inter-stage dependency. The cascaded design yields more favorable results on most metrics, especially emotion accuracy and BERT-F1. This suggests that the gain comes not only from introducing multiple agents, but also from organizing them as a progressive reasoning pipeline. By passing grounded evidence, forecasted emotion, and pragmatic strategy across stages, the framework provides more coherent guidance for downstream generation and thus leads to better final responses.

\begin{table}[t]
\centering
\small
\setlength{\tabcolsep}{3.6pt}
\caption{Ablation study on core agents and reasoning-to-generation pipeline organization in the proposed framework.}
\label{tab:component_analysis}
\begin{tabular}{lcccccc}
\hline
\textbf{Ablation} & \textbf{PPL}$\downarrow$ & \textbf{Dist-1/2}$\uparrow$ & \textbf{Acc.}$\uparrow$ & \textbf{P.}$\uparrow$ & \textbf{R.}$\uparrow$ & \textbf{F.}$\uparrow$ \\
\hline
\multicolumn{7}{l}{\textit{Agent-wise ablation}} \\
w/o A.1 (MPA)             & 2.02 & 34.65/80.23 & 77.42 & 83.90 & 86.43 & 85.12 \\
w/o A.2 (CAEF)            & 1.43  & 32.72/76.09 & --& 83.09 & 86.14 & 84.56 \\
w/o A.3 (PSP)             & \textbf{1.40} & 34.73/75.36 & 80.65 & 83.16 & 86.00 & 84.52 \\
w/o A.5 (GRA)             & 1.44 & 32.20/73.43 & 74.19 & 82.98 & 85.61 & 84.25 \\
\hline
\multicolumn{7}{l}{\textit{pipeline organization}} \\
 Cascaded       & 1.94 & \textbf{37.25/81.10} & \textbf{83.87} & \textbf{84.03} & \textbf{86.64} & \textbf{85.28} \\
Independent    & 1.90 & 36.78/79.03 & 74.19 & 83.28 & 85.55 & 84.37 \\
\hline
\end{tabular}
\end{table}

\begin{table}[t]
\centering
\small
\setlength{\tabcolsep}{3.6pt}
\caption{Effect of maximum iteration count and final response selection in the closed-loop reflection and refinement module.}
\label{tab:tmax_selection}
\begin{tabular}{lccccccc}
\hline
\textbf{Setting} & \textbf{PPL}$\downarrow$ & \textbf{Dist-1/2}$\uparrow$ & \textbf{Acc.}$\uparrow$ & \textbf{P.}$\uparrow$ & \textbf{R.}$\uparrow$ & \textbf{F.}$\uparrow$ & \textbf{Avg. $t^*$} \\
\hline
$T_{\max}=1$ & 1.38 & 29.92/74.39 & 74.19 & 82.92 & 86.28 & 84.53 & 0.26 \\
$T_{\max}=2$ & 1.94 & 37.25/81.10 & \textbf{83.87} & \textbf{84.03} & \textbf{86.64} & \textbf{85.28} & 1.16 \\
$T_{\max}=3$ & \textbf{1.29} & 30.79/73.05 & 74.19 & 82.82 & 85.50 & 84.10 & 1.35 \\
$T_{\max}=4$ & 1.43 & 32.58/74.03 & 83.87 & 82.65 & 85.77 & 84.15 & 1.26 \\
$T_{\max}=5$ & 1.38 & 34.00/77.09 & 74.19 & 82.71 & 85.62 & 84.17 & 1.42 \\
$T_{\max}=6$ & 1.47 & 34.41/72.91 & 77.42 & 83.10 & 85.60 & 84.29 & 1.23 \\
\hline
\multicolumn{8}{l}{\textit{Without final response selection}} \\
$T_{\max}=2$  & 2.02 & \textbf{37.55/81.82} & 77.42 & 83.73 & 85.97 & 84.80 & -- \\
\hline
\end{tabular}
\end{table}

\textbf{Effect of Maximum Iteration Count and Final Response Selection.}
We next analyze two design choices in the closed-loop reflection and refinement module: the maximum iteration count $T_{\max}$ and the final response selection mechanism. As shown in Table \ref{tab:tmax_selection}, among the tested settings, $T_{\max}=2$ provides the most balanced performance. Compared with $T_{\max}=1$, it brings clear gains in diversity and emotion accuracy, while further increasing $T_{\max}$ does not yield consistent improvement. This suggests that a small number of refinement iterations is sufficient to correct major stage-level mismatches, whereas additional iterations may introduce over-correction or semantic drift.

A second observation from Table \ref{tab:tmax_selection} is that effective refinement usually requires only a small number of iterations. As the maximum iteration count increases, the average selected iteration remains concentrated in the early rounds, suggesting that most useful corrections are made early and that additional iterations bring limited benefit. This is further supported by the variant without final response selection, where directly using the last-iteration output leads to lower emotion accuracy and slightly weaker semantic matching.

\textbf{Effect of Retrieval Top-$k$.}
We further analyze the role of the retrieval library by varying the number of retrieved exemplars used during generation. The results indicate that retrieval is helpful when used sparingly. Small top-$k$ settings yield modest gains in semantic matching, while increasing $k$ does not lead to consistent improvement in BERTScore. This suggests that a limited number of relevant exemplars can provide useful auxiliary support, whereas excessive retrieved context may introduce noise.

\begin{table}[t]
\centering
\small
\setlength{\tabcolsep}{3.6pt}
\caption{Effect of retrieval top-$k$ in the auxiliary retrieval library.}
\label{tab:retrieval_analysis}
\begin{tabular}{lcccccc}
\hline
\textbf{Setting} & \textbf{PPL}$\downarrow$ & \textbf{Dist-1/2}$\uparrow$ & \textbf{Acc.}$\uparrow$ & \textbf{P.}$\uparrow$ & \textbf{R.}$\uparrow$ & \textbf{F.}$\uparrow$ \\
\hline
w/o Retrieval & 1.91 & 37.06/80.00 & 74.20 & 83.81 & 86.05 & 84.88 \\
Top-$k$ = 1   & 1.94 & 37.25/81.10 & \textbf{83.87} & 84.03 & \textbf{86.64} & 85.28 \\
Top-$k$ = 2   & 1.91 & 36.13/77.66 & 77.42 & 83.66 & 86.56 & 85.05 \\
Top-$k$ = 3   & \textbf{1.87} & 37.19/79.71 & 80.65 & \textbf{84.40} & \textbf{86.64} & 85.47 \\
Top-$k$ = 4   & \textbf{1.87} & 37.33/80.57 & 77.42 & 83.87 & 86.58 & \textbf{85.17} \\
Top-$k$ = 5   & 1.92 & \textbf{37.57/81.36} & 74.20 & 83.44 & 86.13 & 84.74 \\
\hline
\end{tabular}
\end{table}

\textbf{Framework Performance across Different Backbones.}
We further examine whether the proposed framework remains effective across backbone models of different capacities. Table~\ref{tab:backbone_generalization} shows that the framework remains applicable to both larger and relatively smaller backbones, with smaller models still achieving reasonable performance. Meanwhile, the final results still vary with backbone capability, indicating that stronger models provide more reliable support for multimodal reasoning and response generation.

\begin{table}[t]
\centering
\small
\setlength{\tabcolsep}{3.6pt}
\caption{Framework performance across backbone models of different capacities.}
\label{tab:backbone_generalization}
\begin{tabular}{lcccccc}
\hline
\textbf{Backbone Model} & \textbf{PPL}$\downarrow$ & \textbf{Dist-1/2}$\uparrow$ & \textbf{Acc.}$\uparrow$ & \textbf{P.}$\uparrow$ & \textbf{R.}$\uparrow$ & \textbf{F.}$\uparrow$ \\
\hline
Qwen3.5:27B                & 1.94 & 37.25/81.10 & \textbf{83.87} & 84.03 & \textbf{86.64} & \textbf{85.28} \\
Qwen3.5:9B                 & 2.27 & 38.22/\textbf{85.98} & 80.64 & 83.24 & 85.92 & 84.53 \\
Qwen3-VL:32B     & 2.07 & 38.14/80.73 & 77.42 & 83.63 & 85.96 & 84.75 \\
Qwen2.5-VL:7B              & 1.71 & 32.21/70.41 & 67.74 & 83.84 & 85.78 & 84.77 \\
Gemma3:12B                 & \textbf{1.21} & \textbf{40.19}/83.86 & \textbf{83.87} & \textbf{84.22} & 85.80 & 84.97 \\
LLaVA:13B                  & 1.70 & 18.46/56.99 & 74.19 & 80.38 & 84.32 & 82.27 \\
\hline
\end{tabular}
\end{table}

\subsection{Qualitative Analysis}
We further analyze the internal behavior of the pragmatic strategy planner (PSP) and global reflection agent (GRA) to better understand how our model works beyond aggregate metrics.

\begin{figure}[t]
    \centering
    \includegraphics[width=1\linewidth]{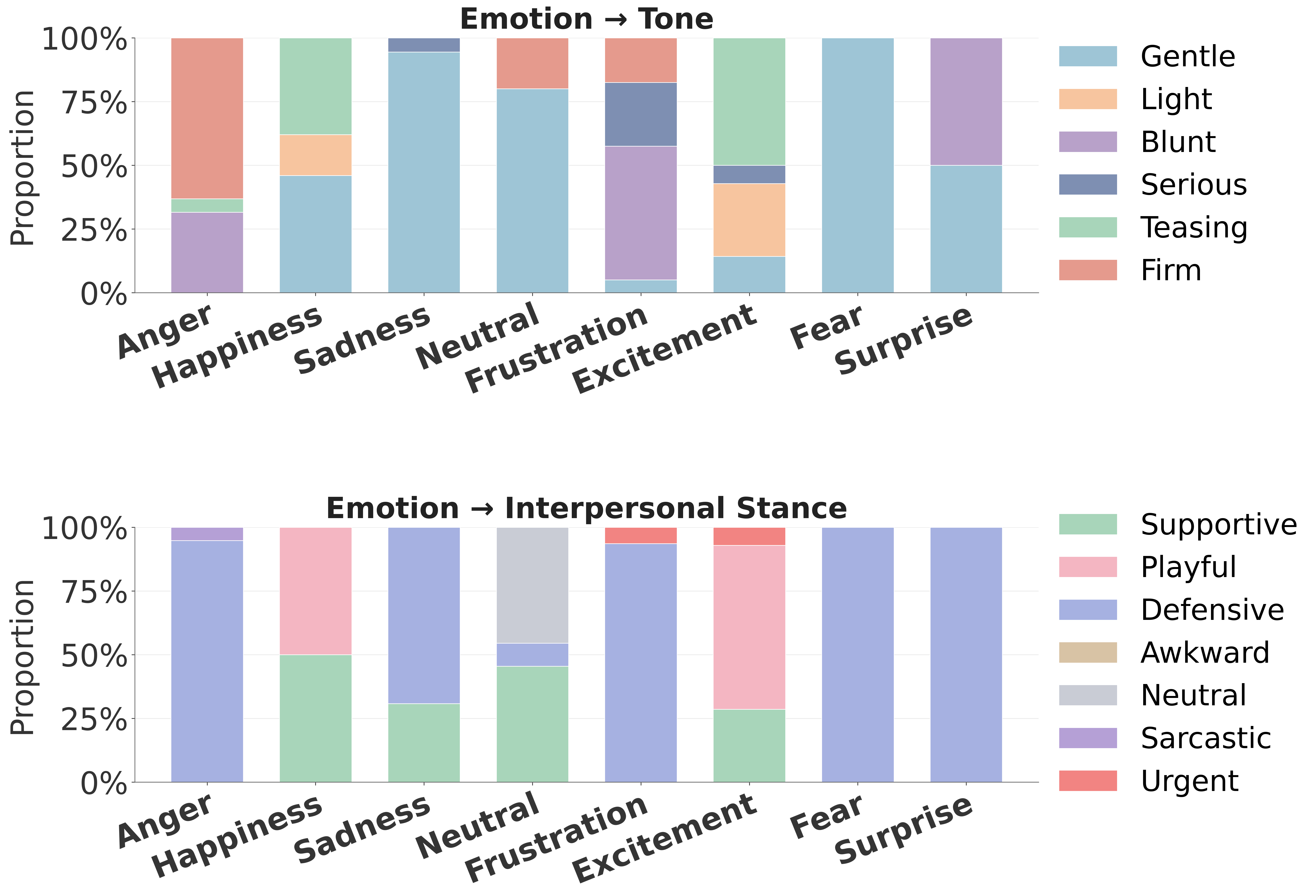}
    \caption{Tone and interpersonal stance distributions under different forecasted emotions.}
    \label{fig:stra}
\end{figure}

\textbf{PSP Strategy Patterns under Different Forecasted Emotions.}
We first examine whether PSP adapts response strategy to different forecasted emotions. As shown in Fig.~\ref{fig:stra}, tone and interpersonal stance exhibit clear emotion-conditioned patterns. Sadness and fear are more often associated with gentler tones, whereas anger and frustration tend to correspond to firmer or blunter strategies. Happiness and excitement show relatively more light, teasing, supportive, or playful patterns. Overall, these results suggest that PSP performs emotion-sensitive pragmatic adaptation rather than using a fixed response style.

\begin{figure}[t]
    \centering
    \includegraphics[width=1\linewidth]{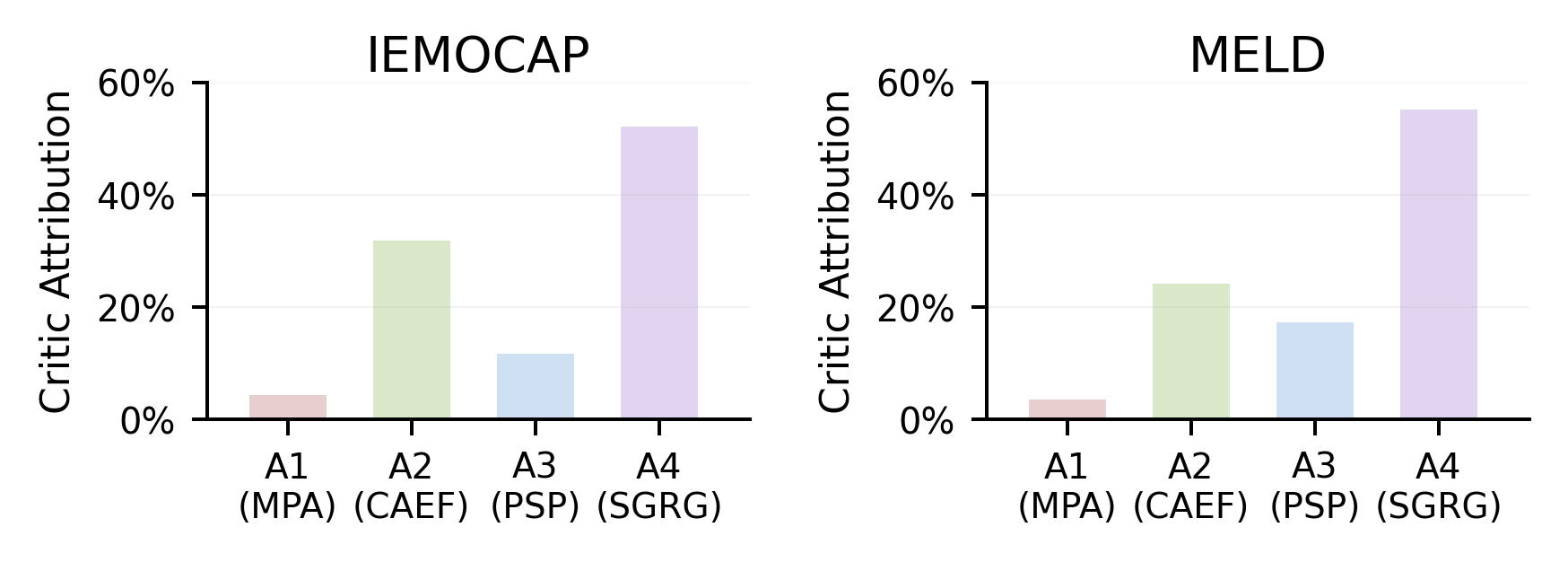}
    \caption{Distribution of GRA error attributions across A1--A4 on IEMOCAP and MELD.}
    \label{fig:critic}
\end{figure}

\textbf{GRA Error Attribution Patterns.}
We next examine whether GRA can globally identify the agent that should be re-run. As shown in Fig.~\ref{fig:critic}, revisions are concentrated on A2 (CAEF) and A4 (SGRG), while A1 (MPA) is selected much less frequently. This suggests that GRA does not revise the pipeline uniformly, but is able to route feedback to the more responsible stage. It also indicates that emotion forecasting and response generation are relatively more challenging stages, and therefore are more likely to require regeneration.

\begin{figure}[ht]
    \centering
    \includegraphics[width=1\linewidth]{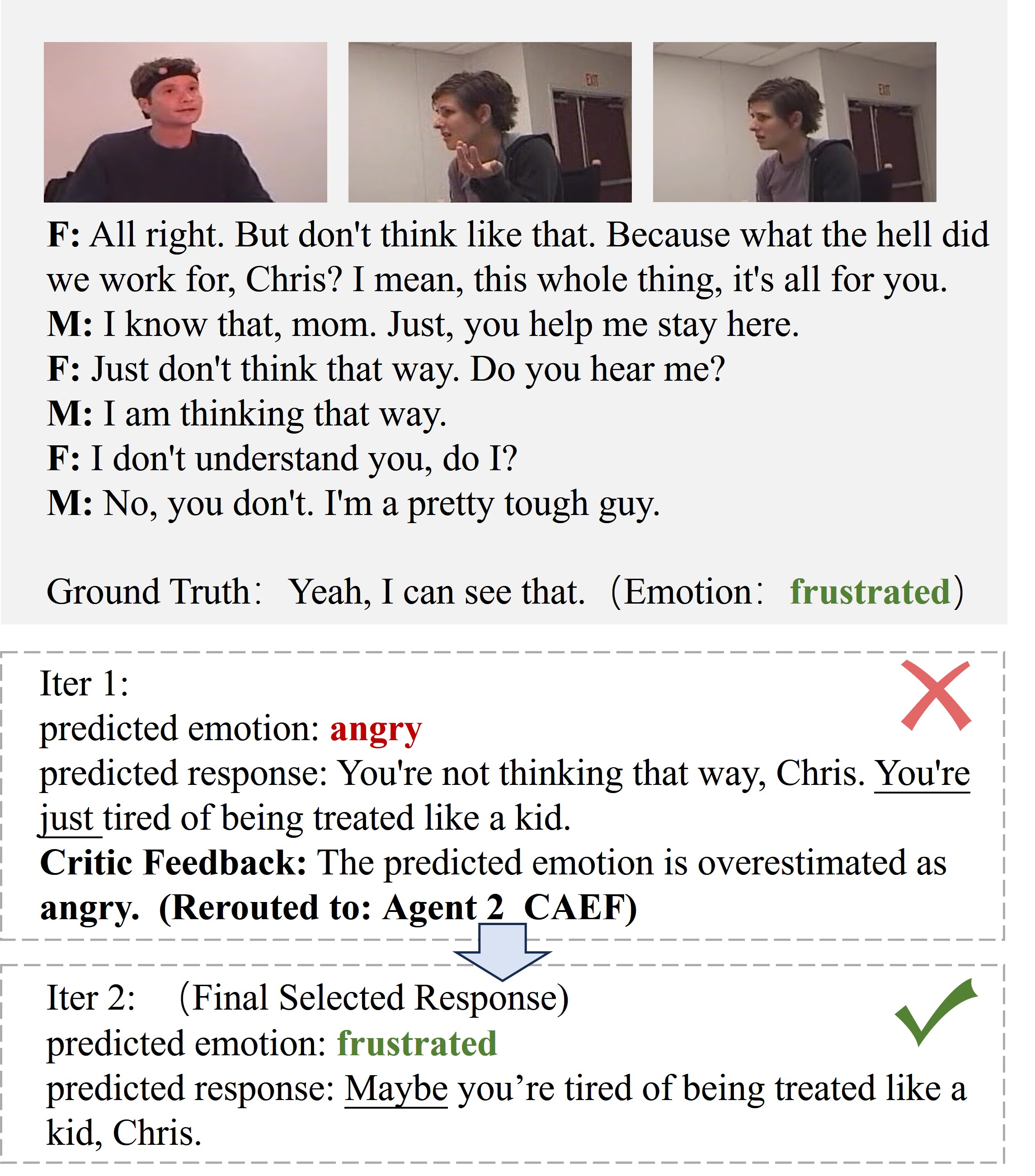}
    \caption{Case study of closed-loop reflection. The critic identifies that the initial emotion prediction (\textit{angry}) is overstated, reroutes refinement to Agent 2, and corrects it to \textit{frustrated}, which further leads to a more appropriate response phrasing.}
    \label{fig:case}
\end{figure}

\textbf{Case Study.}
Fig.~\ref{fig:case} shows a representative example of closed-loop reflection in our framework. In this dialogue, the speaker’s intense expression makes the target emotion easy to overestimate, and Agent 2 (CAEF) initially predicts \textit{angry}. The critic then identifies this prediction as overstated and routes the revision back to Agent 2, where the emotion is corrected to \textit{frustrated}. This illustrates the critic’s ability to locate the source of the error and enable targeted correction.
The example also shows that emotion prediction affects downstream generation. With \textit{angry}, the response uses ``You're just,'' which sounds more direct and assertive; after the emotion is revised to \textit{frustrated}, the final response instead adopts ``Maybe,'' yielding a softer and better calibrated formulation. This highlights the role of the structured empathetic reasoning-to-generation pipeline in shaping the final response through emotion-aware generation.

\section{Conclusion}


In this paper, we propose a multi-agent framework for MERG, which leverages a closed-loop iterative optimization network to overcome the inherent limitations of conventional one-pass generation paradigms, such as inaccurate affective perception and biased response generation. Such closed-loop paradigm not only enhances the precision and interpretability of emotion perception but also facilitates the identification of affective deviations and the refinement of responses through a reflective loop, thereby significantly boosting the system's empathetic capacity. Specifically, we introduce a structured reasoning-to-generation pipeline that decouples the complex task into several cascaded agents, which not only refines the hierarchical emotional perception but also enables fine-grained error localization, allowing discrepancies in the generation process to be precisely traced back to specific functional stages. Furthermore, we introduce a global reflection and refinement module, centered around a global reflection agent that audits each individual agent to pinpoint error origins and guides iterative response corrections. Overall, such a closed-loop framework enables our model to gradually improve the accuracy of emotion perception and eliminate emotion biases during the iteration process. Experiments on several benchmarks, e.g., IEMOCAP and MELD, demonstrate that our model has superior empathic response generation capabilities compared to state-of-the-art methods.

\bibliographystyle{ACM-Reference-Format}
\bibliography{sample-base}

\end{document}